\documentclass[letterpaper]{article}
\usepackage{aaai2026} 
\usepackage{times}
\usepackage{helvet}
\usepackage{courier}
\usepackage[hyphens]{url}
\usepackage{graphicx}
\usepackage{natbib}
\usepackage{caption}
\usepackage{amssymb}
\usepackage{amsmath}
\usepackage{algorithm}
\usepackage{algorithmic}
\usepackage{newfloat}
\usepackage{listings}
\usepackage{xcolor}
\definecolor{lightblue}{rgb}{0.8, 0.85, 1} 
\usepackage[most]{tcolorbox}
\DeclareCaptionStyle{ruled}{labelfont=normalfont,labelsep=colon,strut=off}
\floatstyle{ruled}
\newfloat{listing}{tb}{lst}{}
\floatname{listing}{Listing}

\title{Beyond Wide-Angle Images: Structure-to-Detail Video Portrait Correction via Unsupervised Spatiotemporal Adaptation}
\author{
    Wenbo Nie\textsuperscript{\rm 1,\rm 2},
    Lang Nie\textsuperscript{\rm 3}\footnotemark[2],
    Chunyu Lin\textsuperscript{\rm 1,\rm 2}\footnotemark[1],
    Jingwen Chen\textsuperscript{\rm 1,\rm 2},
    Ke Xing\textsuperscript{\rm 1,\rm 2},
    Jiyuan Wang\textsuperscript{\rm 1,\rm 2},
    Kang Liao\textsuperscript{\rm 4}
}
\affiliations{
    \textsuperscript{\rm 1}Institute of Information Science, Beijing Jiaotong University, Beijing, China\\
    \textsuperscript{\rm 2}Visual Intelligence + X International Joint Laboratory of the Ministry of Education, Beijing, China\\
    \textsuperscript{\rm 3}Chongqing University of Posts and Telecommunications, Chongqing, China\\
    \textsuperscript{\rm 4}Nanyang Technological University, Singapore\\

}

\usepackage{bibentry}

\begin{document}

\twocolumn[{
    \renewcommand\twocolumn[1][]{#1}
    \maketitle
    \vspace{-0.3cm}
    \begin{center}
        \vspace{2mm}
        \centering\captionsetup{type=figure}
        \includegraphics[width=\textwidth]{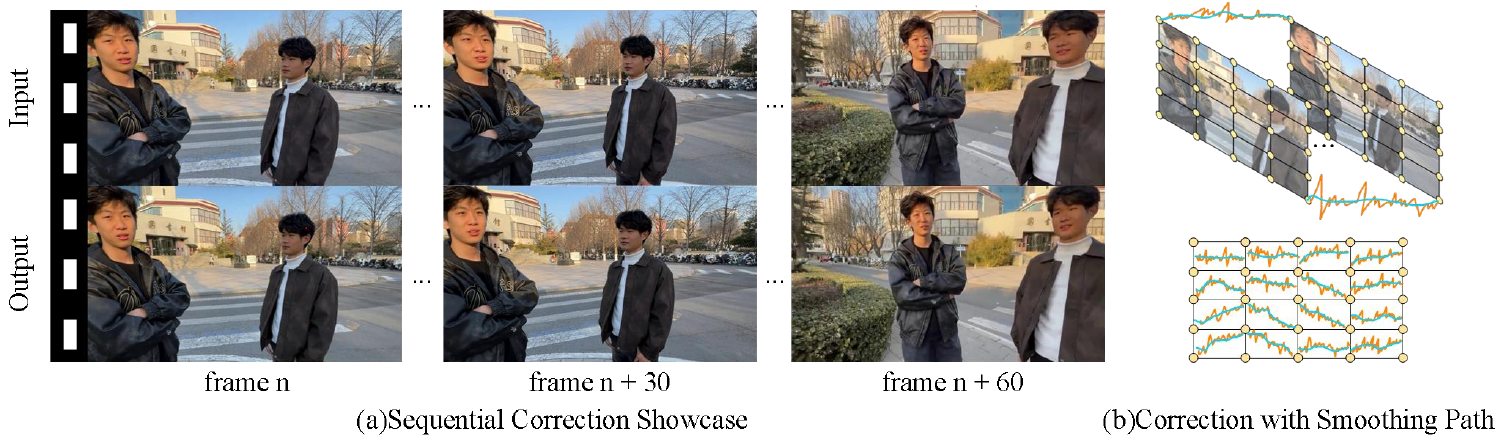}
        \captionof{figure}{Our wide-angle video portrait correction results. (a) The first row shows a wide-angle phone video with edge distortion, corrected in the second row using our method while preserving spatiotemporal consistency. (b) Video correction introduces temporal shake (orange line in the figure), while our method smooths the trajectory (blue line).} 
        \label{fig:cover}
    \end{center}
}]
\begin{abstract}
Wide-angle cameras, despite their popularity for content creation, suffer from distortion-induced facial stretching—especially at the edge of the lens—which degrades visual appeal.
To address this issue, we propose a structure-to-detail portrait correction model named \textbf{ImagePC}.
It integrates the long-range awareness of the transformer and multi-step denoising of diffusion models into a unified framework, achieving global structural robustness and local detail refinement.
Besides, considering the high cost of obtaining video labels, we then repurpose ImagePC for unlabeled wide-angle videos (termed \textbf{VideoPC}), by spatiotemporal diffusion adaption with spatial consistency and temporal smoothness constraints. For the former, we encourage the denoised image to approximate pseudo labels following the wide-angle distortion distribution pattern, while for the latter, we derive rectification trajectories with backward optical flows and smooth them. 
\let\thefootnote\relax\footnotetext{*Corresponding author; \dag~project lead}
Compared with ImagePC, VideoPC maintains high-quality facial corrections in space and mitigates the potential temporal shakes sequentially in \textit{blind scenarios}.
Finally, to establish an evaluation benchmark and train the framework, we establish a video portrait dataset with a large diversity in the number of people, lighting conditions, and background.
Experiments demonstrate that the proposed methods outperform existing solutions quantitatively and qualitatively, contributing to high-fidelity wide-angle videos with stable and natural portraits. The codes and dataset will be available. The video is available at
\textcolor{blue!60!white}{\url{https://wenbo-nie.github.io/structure-to-detail-portrait-correction/}}.

\end{abstract}

\section{Introduction}
With the development of the self-media and videography industries, wide-angle lenses have become increasingly favored for the capability of capturing expansive scenes. However, such lenses inevitably introduce geometric distortions, particularly pronounced at the lens boundaries, resulting in background straight line curvatures and facial feature deformations in still images and video recordings.
    
Traditional wide-angle portrait correction methods typically require precise camera parameters (\textit{e.g.}, focal length \cite{10.1145/3306346.3322948}) as a prerequisite, followed by stereographic projection, face and line detection, and energy optimization to achieve geometric correction \cite{10.1145/3306346.3322948}. Such a pipeline is complex in process and inapplicable when the relevant camera parameters are unknown. 
The same goes for video portrait correction work~\cite{Lai_2022}.
   In contrast, learning-based solutions eliminate this issue~\cite{tan2021practicalwideangleportraitscorrection} by directly learning the spatial mapping from wide-angle images to rectification flows in a supervised manner. However, these methods often exhibit noticeable temporal shakes when applied to videos. Moreover, there is currently a lack of datasets for video portrait correction due to the significantly higher cost of labeling video data.

    To this end, we take a pioneering step in achieving video portrait correction without video annotations. First of all, we present a structure-to-detail portrait correction model, named \textbf{ImagePC}, to generate high-quality rectification flows, which integrates the advantages of both Transformer and diffusion models.
    The Transformer model establishes long-range geometric dependencies by capturing spatial global features across entire images, providing structural guidance for the subsequent diffusion process.
    The diffusion model refines high-fidelity flow patterns through iterative denoising guided by Transformer-derived structural information. This hybrid architecture enables structure-to-detail pixel-wise correction, where structural priors from the Transformer and detail-generation capacity from the diffusion model are synergistically combined. 
    
    More importantly, we propose an unsupervised spatiotemporal adaption approach to enable ImagePC the capability to rectify video sequences stably. The repurposed model, which we term \textbf{VideoPC}, addresses the issue of temporal shakes without requiring video correction labels. 
    Concretely, we first use the pre-trained ImagePC to generate rectification flows for each video frame as pseudo-labels. When ImagePC is adapted to unseen wide-angle videos, these pseudo-labels can guide the denoising process, ensuring spatial consistency in facial corrections with the wide-angle distortion distribution prior.
    To enforce temporal smoothness, we derive rectification trajectories across sequential frames using backward optical flows and design a temporal smoothness constraint to mitigate sequential jitters. 
    With these spatiotemporal constraints, VideoPC achieves stable wide-angle video correction in blind scenarios that not only preserves high-fidelity facial details but also frees from labor-intensive video annotations.
    
    To establish an evaluation benchmark and train the proposed video portrait correction model, we construct a wide-angle video dataset with a wide diversity in scene, camera, and number of people.
    Finally, we conduct extensive experiments about portrait correction in both image and video, demonstrating our superiority over other solutions. The contributions center around:
\vspace{5pt}
\begin{itemize}
    \item We propose a structure-to-detail portrait correction model, named \textbf{ImagePC}, which integrates long-range dependences of transformer and iterative refinements of diffusion models for global-to-local rectification.
    \item We design an unsupervised spatiotemporal adaptation framework to transfer \textbf{ImagePC} to \textbf{VideoPC} without labeled videos. It combines spatio-temporal optical flows to track correction trajectories, establishing spatial consistency and temporal smoothness constraints.
    \item We present a video portrait dataset with wide-angle cameras and conduct extensive experiments to validate our superior performance over existing solutions.
\end{itemize}
\section{Related Work}
\label{sec:related_work}
\subsection{Portrait Correction}
    Portrait correction in wide-angle images is a significant research area. Traditional methods, such as Shih et al.'s\cite{10.1145/3306346.3322948} content-aware warping, adjust distortions by analyzing image content and camera parameters. However, these methods 
   often rely on camera parameters or complex inputs and are difficult to extend to video processing. Stereographic projection methods\cite{10.1145/3306346.3322948} can preserve local conformality but struggle with temporal consistency in videos.

    Deep learning has introduced calibration-free solutions. Tan et al.\cite{tan2021practicalwideangleportraitscorrection} proposed a two-stage neural network leveraging wide-angle images and correction flow datasets. In contrast, Zhu et al.~\cite{zhu2022semisupervisedwideangleportraitscorrection} developed a semi-supervised Transformer to reduce annotation costs. Nie et al.’s method CoupledTPS\cite{Nie_2024}, iteratively couples limited TPS models with a warping flow to precisely correct facial features. 

    Existing deep learning methods, designed for static images, fail to address the temporal dynamics of videos due to expensive frame-by-frame annotation and limited datasets.
    Lai et al.\cite{Lai_2022} proposed a temporal consistency optimization method using spherical projection and energy minimization, but it struggles to integrate global video information with facial correction details and lacks adaptability in complex environments. 
    To address this issue, we propose learning video correction principles from image networks.
 \subsection{Video stabilization}
Traditional video stabilization techniques primarily stabilize the video by smoothing the feature trajectories between multiple frames or adjacent frames~\cite{6420828,5459297,6126544}. Subsequently, grid-based methods~\cite{6420828,10.1145/2461912.2461995,Liu2016MeshFlowML,unyan2024} and optical flow  ~\cite{Liu_2014_CVPR}; ~\cite{james2022globalflownetvideostabilizationusing} have also been applied to represent motion, becoming key tools for stabilizing videos. With the rapid development of deep learning, deep learning methods have been applied to video stabilization technology~\cite{8951447,8953555,zhao2023fastfullframevideostabilization,peng20243dmultiframefusionvideo,4106837}, where the original video frames are input, and stabilized video frames are directly output. For example, Wang et al. ~\cite{8554287} proposed an end-to-end learning framework, and ~\cite{https://doi.org/10.1111/cgf.13566} used adversarial networks to generate target images. DUT ~\cite{9801555}, which learns video stabilization by simply watching unstable videos. Additional methods, including unsupervised online video stitching ~\cite{nie2024eliminatingwarpingshakesunsupervised}, and meta-learning-based approaches to improve full-frame stabilization
  ~\cite{ali2024harnessingmetalearningimprovingfullframe}, have been explored to enhance stability. These approaches, ranging from optical flow to deep learning models, provide valuable solutions for addressing video temporal shake in video portrait correction.

\subsection{Diffusion Models}
    Diffusion models ~\cite{yang2024diffusionmodelscomprehensivesurvey} excel in generative tasks, extending from image generation to video-related applications. In video generation, early advancements include Video Diffusion Models (VDM)~\cite{ho2022videodiffusionmodels}, which adapted the 2D U-Net architecture from image DMs to a 3D U-Net for generating coherent video frames. To enhance temporal resolution, Ho et al.\cite{ho2022imagenvideohighdefinition}  introduced a cascade of video super-resolution models, improving the quality of generated sequences. Additionally, methods like Make-A-Video~\cite{singer2022makeavideotexttovideogenerationtextvideo} achieved text-to-video generation by leveraging pre-trained text-to-image DMs, avoiding the need for paired text-video data.

    More recent work focuses on video editing and enhancing temporal consistency. For example, Tune-A-Video~\cite{wu2023tuneavideooneshottuningimage} fine-tunes models on  videos for coherent outputs, while FateZero~\cite{qi2023fatezerofusingattentionszeroshot} ensures temporal consistency using attention-based blending techniques. Similarly, Rerender-A-Video~\cite{yang2023rerendervideozeroshottextguided} and VideoControlNet\cite{hu2023videocontrolnetmotionguidedvideotovideotranslation}  introduce pixel-aware latent fusion and motion-aware adjustments to maintain visual consistency during editing.

\begin{figure*}[t!]
  \centering
  \includegraphics[width=\textwidth]{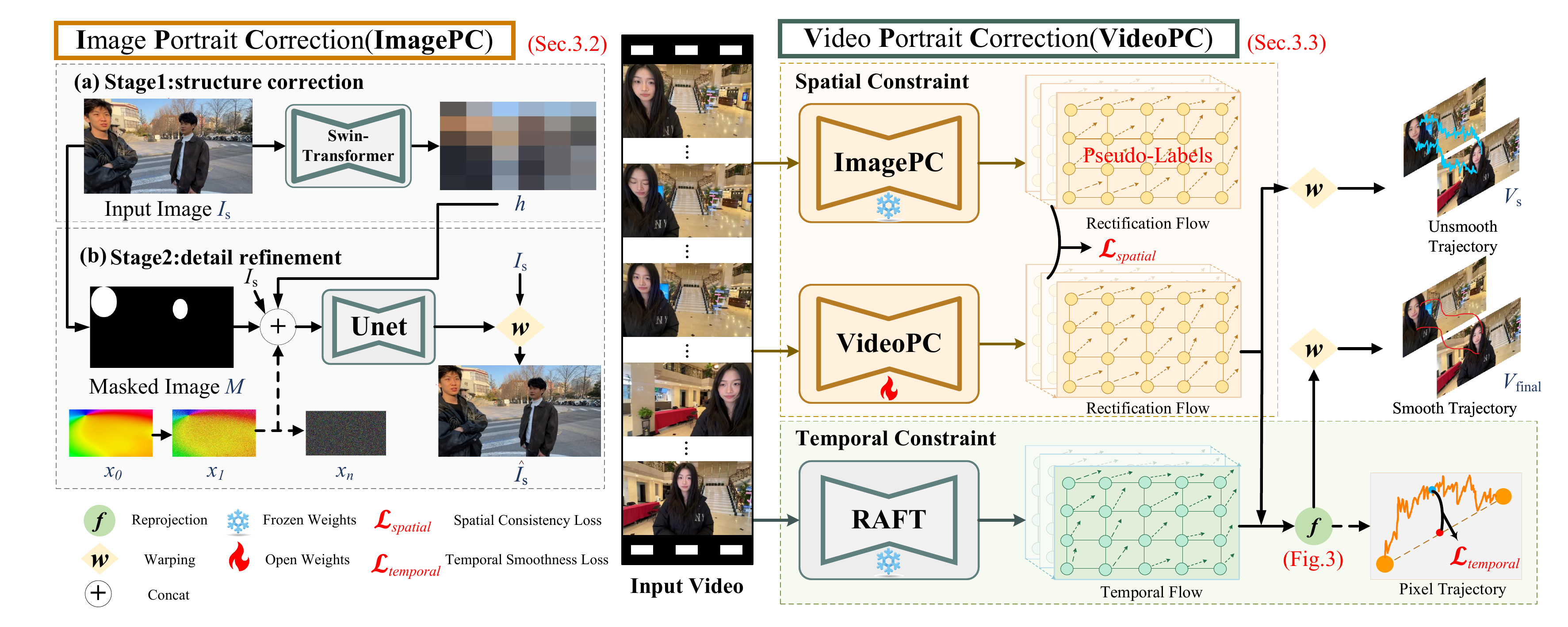}
  \caption{The proposed framework consists of two models. (1) ImagePC: We design a structure-to-detail architecture to generate high-quality rectification flows for each individual frame. (2) VideoPC: We adapt ImagePC from the image to video domain with spatial consistency and temporal smoothness constraints.}
  \label{fig:framework}
  \vspace{-0.3cm}
\end{figure*}

\vspace{-1pt}

\section{Methodology}
\label{sec:methodology}
\subsection{Overview}


    The overview of our framework is shown in Fig.~\ref{fig:framework}, consisting of an image correction model and a video correction model.
    In the image model, ImagePC tackles the distorted image $I_s$, producing precise optical flows using a structure-to-detail rectification architecture. The predicted flow is then applied to generate a geometrically corrected image $\hat{I_s}$. 

In the video model, directly adopting the image model to correct the video frames can easily yield temporal shakes due to temporally unsmooth warps. But we could take the correction flows from the image model as the reference and transfer the image model to the video domain through spatiotemporal adaptation.


\subsection{ImagePC}

    As illustrated in Fig.~\ref{fig:framework}(a,b), our ImagePC is a framework for high-quality portrait correction by predicting precise optical flow fields to rectify facial distortions. 
    \subsubsection{Structure-to-Detail Model}
    In the first stage of ImagePC, we focus on capturing the global geometric structure of the distorted input image $I_s$. The Transformer network \cite{zhu2022semisupervisedwideangleportraitscorrection}, in Fig.~\ref{fig:framework}(a), with its advantage in capturing long-range information, extracts a set of intermediate layer features $h$ that effectively reflect the global geometric structure of the distorted input image. 
    These features $h$ serve as a structural prior, guiding the diffusion model in the next stage to accurately align the facial geometry while preserving spatial coherence.

    Building on the global structure from the first stage, the next stage, as shown in Fig.~\ref{fig:framework}(b), refines local details using a diffusion-based approach. To ensure that the correction process primarily focuses on the area of the face, we use InsightFace \cite{ren2023pbidr,guo2021sample,unGuo2024} to process the original facial image $I_{\text{s}}$, allowing accurate face detection and keypoint localization. Based on these detection results, a binary mask \textit{M} is generated, which clearly delineates the facial region to enable better portrait perception. This mask provides spatial guidance for subsequent processing and improves the quality of the correction.
    Leveraging the detail-handling capability of diffusion models, we form the conditioning input for ImagePC by concatenating $M$, $I_s$, and $h$:
    \begin{equation}
        x_{in}=cat[M, I_{s}, h].
    \end{equation}
   Our approach leverages the iterative refinement capabilities of Denoising Diffusion Implicit Models (DDIM)\cite{song2022denoisingdiffusionimplicitmodels}, enabling iterative data generation through noise-adding and denoising processes. 

    Adopting this concept, ImagePC redefines the diffusion process to operate directly on optical flow fields.

    Unlike traditional image-to-image methods that directly generate a corrected facial image, we modify the diffusion network in this stage to predict an optical flow field $ F^{corr} $. The repurposed image-to-flow pipeline not only ensures that the corrected content totally loyal to the original content, but also enables the derivation of video correction trajectories. 
    
    With high-quality correction flows, the wide-angle rectification can be formulated as:
    \begin{equation}
            \hat{I_{s}} =\mathcal{W } (F^{corr},I_{s}).
    \end{equation}
    The warping operation $\mathcal{W}$ realigns the pixels in $I_s$ according to the predicted flow, resulting in a natural facial appearance.
    
    \subsubsection{Objective Function}
    The model is optimized using a comprehensive loss function:
    \begin{equation}
\mathcal{L}_{image}= \lambda_1 \mathcal{L}_{mask}+ \lambda_2 \mathcal{L}_{photo}+ \lambda_3 \mathcal{L}_{flow} ,
\end{equation}
where $\mathcal{L}_{flow}$ ensures optical flow accuracy, $\mathcal{L}_{photo}$ enforces image quality, and $\mathcal{L}_{mask}$ enhances facial edge details, with $\lambda_1$, $\lambda_2$, and $\lambda_3$ as weighting coefficients.

Optical Flow Loss \(\mathcal{L}_{flow}\): It directly measures the motion-level difference between the predicted optical flow and the ground truth optical flow as:
\begin{equation}
    \mathcal{L}_{flow} = \frac{1}{HW} \sum_{h=1}^{H} \sum_{w=1}^{W} W \left( F^{Corr}(h,w) - F_{\text{gt}}(h,w) \right)^2.
    \label{eq:loss_flow}
\end{equation}
Image Loss \(\mathcal{L}_{photo}\): It evaluates the image-level difference between the corrected image and the ground truth image as:
\begin{equation}
\begin{aligned}
    \mathcal{L}_{photo} = \frac{1}{HW} \sum_{h=1}^{H} \sum_{w=1}^{W}W
         \left( \hat{I}_{s}(h,w) - I_{\text{gt}}(h,w) \right)^2.
\end{aligned}
\end{equation}
Mask-based Sobel Loss \(\mathcal{L}_{mask}\): We use the Sobel kernels ($G_x$ and $G_y$) to extract edge information and calculate the edge loss within the facial region mask as:
\begin{equation}
\begin{aligned}
    \mathcal{L}_{mask} &= \left[ \left| G_x(F^{Corr}) - G_x(F_{\text{gt}}) \right| \right. \\
        &\quad + \left| G_y(F^{Corr}) - G_y(F_{\text{gt}}) \right| \big] \cdot M.
    \label{eq:loss_mask}
\end{aligned}
\end{equation}
\subsection{VideoPC}
Videos are dynamic sequences composed of continuous multi-frame images $\mathcal{I} = \left\{ \mathcal{I}_1, \mathcal{I}_2, \dots, \mathcal{I}_n \right\} $. To achieve unsupervised wide-angle video correction, we use pre-trained ImagePC to generate corrected optical flow for each frame as pseudo-labels. However, directly applying these pseudo-labels to the video sequences will result in severe temporal shakes, significantly decreasing the video's quality and visual experience. To address this issue, we further propose to transfer ImagePC to VideoPC with the adaptability in wide-angle video scenarios. 
\subsubsection{Correction Trajectory}
Our method aims to control the comprehensive motion between
consecutive frames.
To represent such motions, we derive correction trajectories that are built on inter-frame and intra-frame optical flows.
For inter-frame motions, we use the RAFT \cite{zhang2024raftadaptinglanguagemodel} to capture the temporal flows ($f_{t \to t+1}$ and $f_{t+1 \to t}$) from wide-angle video frames ($I_t$ and $I_{t+1}$). 
For intra-frame motions, we leverage the VideoPC model (the same architecture as ImagePC) to predict the rectification flows ($F^{corr}_t$). The remaining key question is how to combine these flows into our desired correction trajectories.

\subsubsection{Based on Forward Optical Flow}

For forward optical flow (\textit{i.e.}, $F^{corr}_t$ denotes the correspondences from wide-angle images to rectified images), we can identify the differences of corresponding points within the consecutive rectified results as the following formula:
\begin{equation}
    r(t+1)=f_{t+1 \to t} +F_t^{\mathrm{Corr}} \Bigl( G +f_{t+1 \to t}  \Bigr) - F_{t+1}^{\mathrm{Corr}},
    \label{eqn:forward}
\end{equation}
where $G\in \mathbb{R}^{2\times H\times W}$ denotes pixel-wise grid coordinates. The latter part is the position in the (t+1)-th frame (\textit{i.e.}, $F_{t+1}^{\mathrm{Corr}} + G$), while the upper is the corresponding position in the \textit{t}-th frame (\textit{i.e.}, $f_{t+1 \to t} +F_t^{\mathrm{Corr}}( G +f_{t+1 \to t}) + G$).

\subsubsection{Based on Backward Optical Flow}
However, in the actual portrait distortion correction process, employing forward optical flow to warp is prone to generating invalid spatial holes. 
Therefore, all current optical flow-based correction methods employ backward optical flows, that is, $F^{corr}_t$ denotes the mapping flow from rectified results to original wide-angle images.
The difference in optical flow direction makes the relative motion described in Eq. \ref{eqn:forward} unsuitable for practical backward optical flow transformations.
To compare the rectified results of the $ t $-th and $ (t+1) $-th frames, we need to identify the differences in corresponding positions in the input image for the same grid location. However, directly comparing  the two optical flows \(F^{corr}_t\) and \(F^{corr}_{t+1}\)  is not feasible, as the two backward optical flows do not correspond to the same points. Therefore, we need to leverage \(f_{t\to t+1}\) to correct the temporal misalignment.

\begin{figure}[!t]
  \centering
  \includegraphics[width=0.47\textwidth]{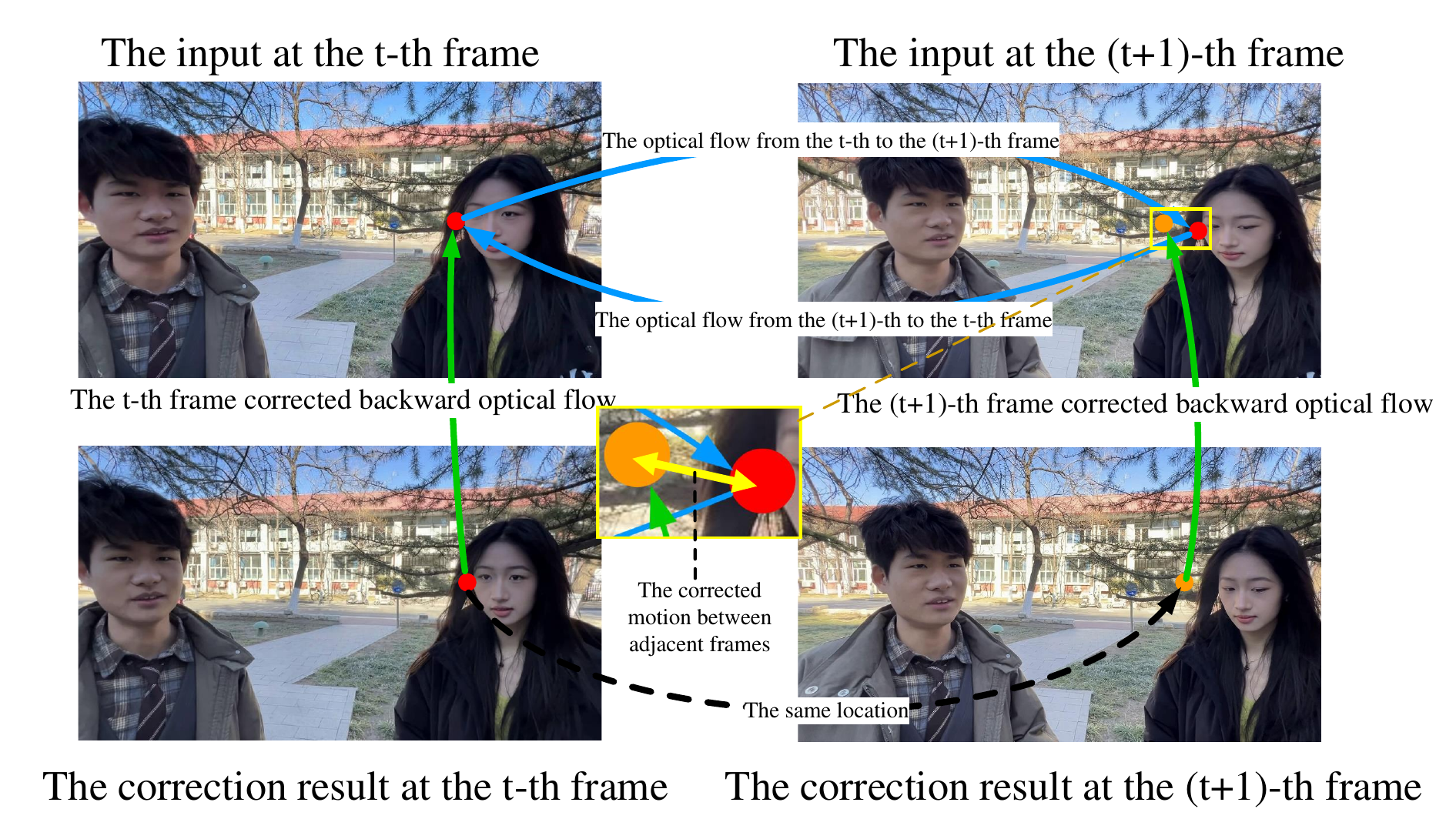}
  \vspace{-3pt}
  \caption{Illustration of the computation of \( r(t+1) \) in Eq.~\ref{eqn:3}, depicting the integration of RAFT optical flow \( f_{t \to t+1} \), corrected mesh \( G \), and correction flows \( F_t^{\mathrm{Corr}} \) and \( F_{t+1}^{\mathrm{Corr}} \) for temporal stabilization. }
  \label{fig:figure3}
  \vspace{-0.5cm}
\end{figure}

The position difference with backward flows is derived as follows:
\begin{equation}
    r(t+1)=f_{t \to t+1} \Bigl( G + F_t^{\mathrm{Corr}} \Bigr) + F_t^{\mathrm{Corr}}  - F_{t+1}^{\mathrm{Corr}},
    \label{eqn:3}
\end{equation}
where Fig. \ref{fig:figure3} provide a visual explanation about Eq. \ref{eqn:3}.

It can be further concatenated in temporal order from the initial moment to obtain the position at each moment as follows:
\begin{equation}
    R(t)=r(1)+r(2)+\cdots+r(t),
\end{equation}
where \( r(1) \) is defined as a zero matrix. The final correction trajectory can be obtained by sequentially chaining \( R(1), R(2), \ldots, R(t), \ldots, R(N) \) over time.

The produced trajectory contains comprehensive motions, including the shakes caused by unsmooth rectification flows and the inherent jitters in original wide-angle videos.

\subsubsection{Spatiotemporal Adaptation}
To achieve unsupervised video correction, we design our objective function $\mathcal{L}_{video}$ concerning two aspects: spatial consistency and temporal smoothness.
\begin{equation}
    \mathcal{L}_{video}=\mathcal{L}_{spatial}+\lambda \mathcal{L}_{temporal}.
\end{equation}

For the temporal smoothness, we encourage the correction trajectory from VideoPC to be as smooth as possible.
 In particular, for three consecutive frames, we require the trajectory position of the middle frame to lie between the trajectory positions of the preceding and following frames: 
    \begin{equation} 
   \mathcal{L}_{temporal} = \| R(t+1) + R(t-1) - 2R(t) \|.
    \end{equation}
    
For spatial consistency, we leverage the pre-trained ImagePC model to ensure the quality of facial distortion correction within each frame. 
Concretely, we use pseudo-labels generated by ImagePC to provide spatial supervision for VideoPC, including the optical flow constraint $\mathcal{L}_{flow}$ (Eq.~\ref{eq:loss_flow}) and mask-based edge constraint $\mathcal{L}_{mask}$(Eq.~\ref{eq:loss_mask}).
In the transfer process of facial correction capabilities from the image to video domain, these components prevent the structural integrity and edge details of the facial regions from severe degradation.
\subsection{Data preparation}
\label{sec:data preparation}
\begin{figure}[t!]
    \centering
    \includegraphics[width=0.97\linewidth]{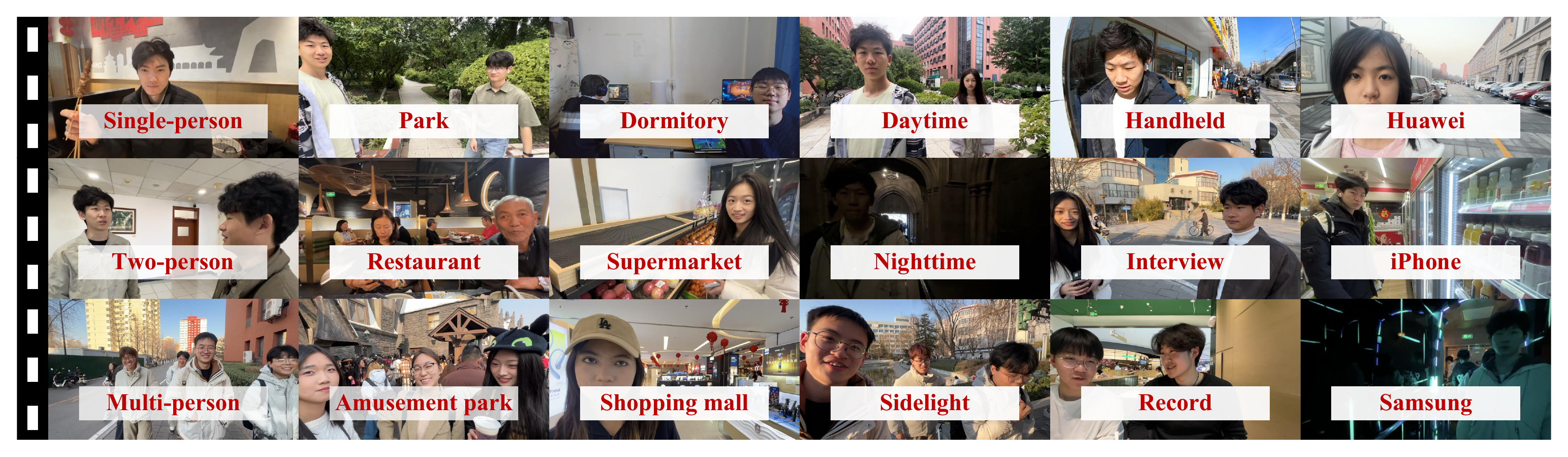}
    \caption{Our wide-angle video face dataset encompasses a broad range of scenes and showcases significant diversity in individuals. We present some examples from our dataset.}
    \label{fig:dataset}
    \vspace{-0.3cm}
\end{figure}

Due to the lack of publicly available wide-angle video portrait datasets, we introduce a new dataset containing 136 clips, each with a resolution of 1080p at 30fps, ranging from 5 seconds to 90 seconds in duration (totaling 43,903 frames). Our dataset includes clips captured by seven smartphones (\textit{e.g.}, iPhone 13 Pro) covering a wide variety of scenarios 
and encompass diverse scenarios, including content captured by self-media creators and film clips extracted from cinematic works. Detailed information is available in the supplementary materials.
To authentically replicate real-world recording conditions, we deliberately introduced artificial shaking to simulate the instability inherent in handheld video capture. This design choice poses significant challenges for stabilization algorithms. Fig.~\ref{fig:dataset} presents some examples of our wide-angle video portrait dataset.
\section{Experiment}
\label{sec:experiment}

\subsection{Dataset and Implementation Detail}

Our methodology builds upon two core datasets: the established wide-angle image dataset~\cite{tan2021practicalwideangleportraitscorrection} and our newly captured video dataset.
\paragraph{Image Data}In our experiments, we use the dataset from Tan et al.\cite{tan2021practicalwideangleportraitscorrection}, which consists of 5133 training images and 129 test images captured with five wide-angle smartphones.
\vspace*{-0.2cm}
\paragraph{Video Data} For both training and testing phases, we utilized the dataset described in Sec.~\ref{sec:data preparation}. The training set comprised 16 video sequences (7,134 frames) that comprehensively represent diverse capture devices, environmental conditions, and scene types, ensuring the model training across varied scenarios.
\begin{figure*}[!h]
  \centering
  \includegraphics[width=0.97\textwidth]{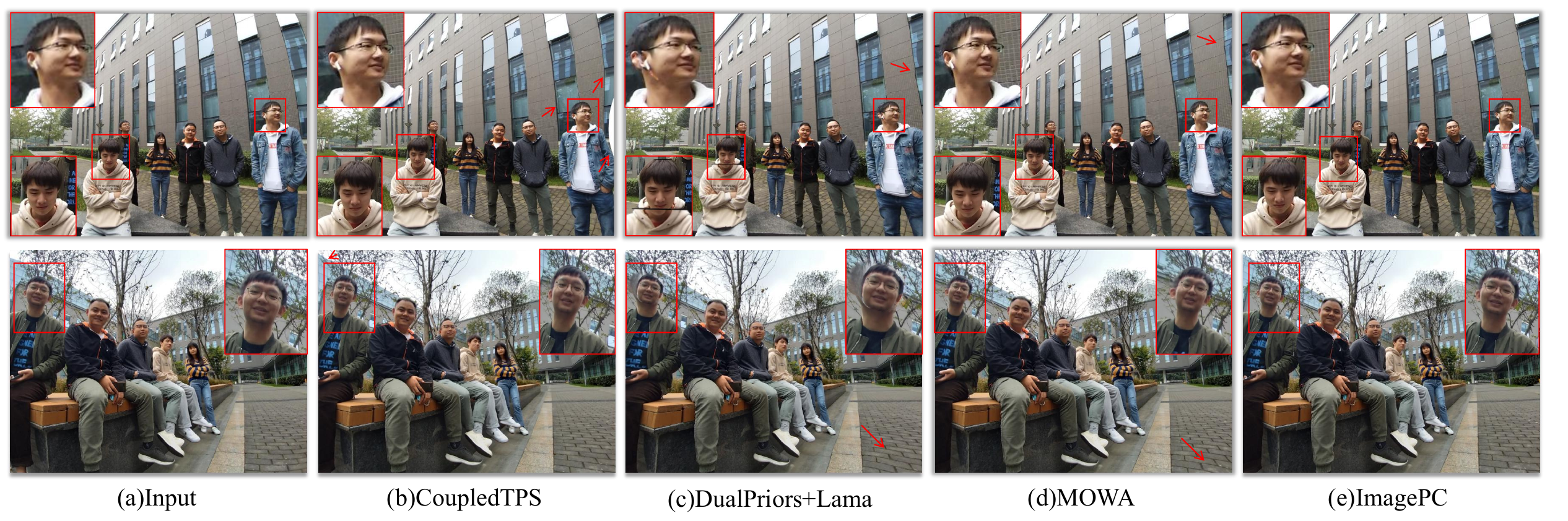}
  \caption{Comparison of ImagePC with existing methods on wide-angle distorted images. ImagePC achieves superior restoration of natural facial features and reduced edge distortions compared to state-of-the-art alternatives.}
  \label{fig:ImagePC}
  \vspace{-0.3cm}
\end{figure*}
\paragraph{Training Process of ImagePC}
Our framework employs a two-stage training protocol. In the first stage, we utilize the Adam optimizer\cite{kingma2017adammethodstochasticoptimization} with an initial learning rate of $1 \times 10^{-4}$, training the model for 200 epochs. The subsequent texture refinement stage leverages high-dimensional features extracted from the initial phase. This feature, combined with image $I_s$ and mask $M$, acts as the conditional guidance for the diffusion model to enhance fine-grained texture details. This stage implements a learning rate of $2 \times 10^{-4}$, with optimization conducted over 500,000 iterations.

\paragraph{Training Process of VideoPC}
For VideoPC, we fine-tune the ImagePC model to maintain correction efficacy while smoothing temporal jitter. Training employs the Adam optimizer\cite{kingma2017adammethodstochasticoptimization} with a learning rate of $2 \times 10^{-4}$ over 300,000 iterations. 
The entire training process is conducted on a \textbf{single RTX 4090}.

\subsection{Evaluation Metric}
Following Tan et al.'s work ~\cite{tan2021practicalwideangleportraitscorrection}, we evaluate geometric correction performance using two established metrics: \textbf{LineAcc} quantifies how well straight edges retain geometric integrity by computing deviations between local slopes and the ideal global orientation. \textbf{ShapeAcc} assesses shape preservation via cosine similarity between reference and corrected landmarks, where higher values indicate better alignment. 
We also employ Jin et al.’s \textbf{Stability Score} (Avg, Trans, Rot)\cite{Choi_TOG20}, which uses homography and FFT, to measure VideoPC’s shake reduction, with higher scores denoting smoother videos.
\subsection{Comparative Result}
We evaluate our approach against existing methods across image and video modalities, providing comprehensive quantitative and qualitative results.

\begin{table}[ht]
\centering
\begin{tabular}{c|c|cc} 
\hline
    Method & Reference & LineAcc & ShapeAcc \\ \hline
    
    Shih et al. & TOG2019 &66.143 & 97.253 \\
    Tan et al.& CVPR2021&66.784 & 97.490 \\
    Zhu et al. &CVPR2022 &66.825 & 97.491 \\
     DualPriors & ECCV2024 & 67.304 & 99.012 \\
    CoupledTPS & TPAMI2024 & 66.808 & 97.500 \\
     MOWA & TPAMI2025 & - & 97.475\\
    ImagePC (Ours) & - &66.898 & 97.508 \\ \hline
    \end{tabular}
    \caption{Quantitative comparison of the proposed ImgaePC with other portrait correction methods. Higher LineAcc and ShapeAcc indicate better straight-line and face correction performance, respectively.}
 \label{tab:image}
\vspace*{-0.5cm}
\end{table}

\textbf{Image Comparison} As shown in Tab. ~\ref{tab:image}, our ImagePC achieves the second-best performance across LineAcc and ShapeAcc metrics, surpassed only by DualPriors\cite{yao2024combininggenerativegeometrypriors}.  This shows that our image correction model can retain the geometric integrity of straight edges and preserve facial shape well. Furthermore, in the qualitative comparison presented in Fig.~\ref{fig:ImagePC}, we can further observe that ImagePC outperforms DualPriors\cite{yao2024combininggenerativegeometrypriors} in the aforementioned aspects.
This undoubtedly demonstrates that, despite DualPriors\cite{yao2024combininggenerativegeometrypriors} achieving state-of-the-art performance on certain metrics, their approach exhibits noticeable abrupt artifacts at stitching boundaries and inconsistent head-to-body proportions in the images. This may be attributed to their method relying on a parallel network that separately rectifies the background and portrait before stitching them into a single image, with blank areas repaired using LaMa\cite{suvorov2021resolution}. In contrast, Fig.~\ref{fig:ImagePC}(c) clearly illustrates that our method generates more natural and refined character appearances, highlighting the superiority of ImagePC.

\begin{figure*}
    \centering
    \includegraphics[width=\textwidth]{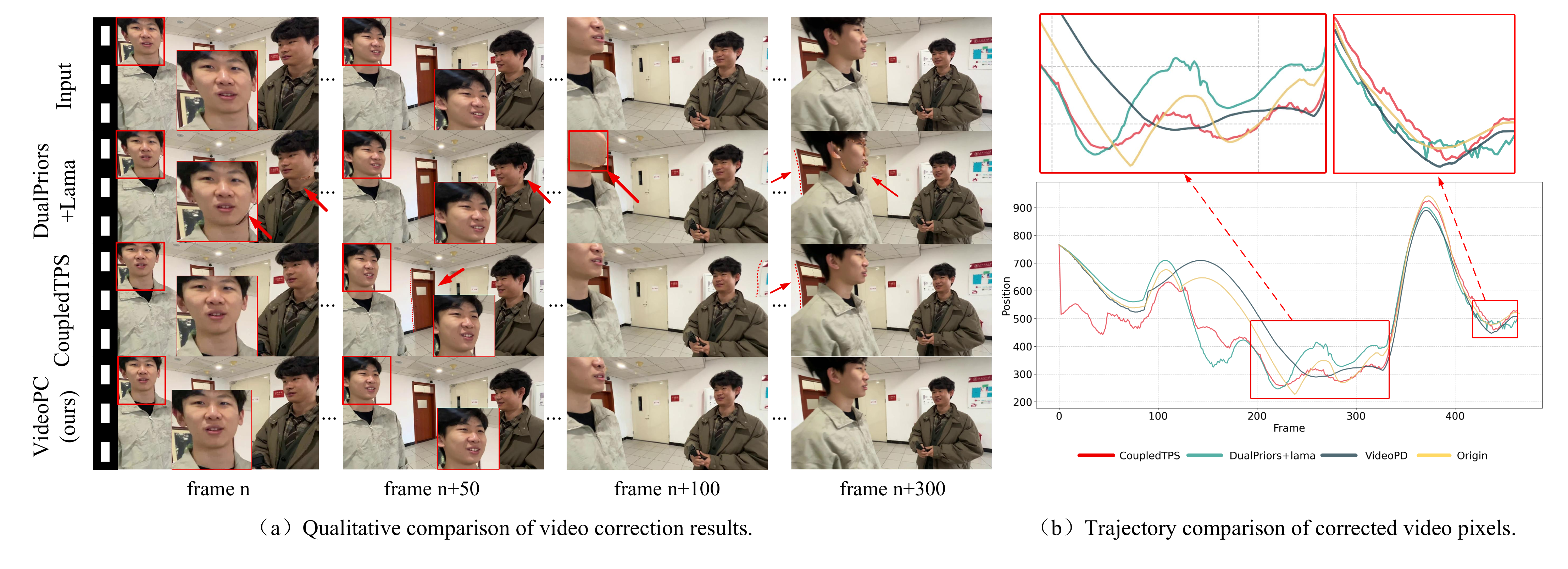}
    \vspace{-0.3cm}
    \caption{Comparison with existing portrait correction methods in wide-angle videos.}
    \label{fig:VideoPC}
    \vspace{-0.3cm}
\end{figure*}

\textbf{Video Comparison}  In the video domain, our approach not only corrects individual frames but also prioritizes inter-frame continuity to mitigate severe visual jitter. As demonstrated in Table~\ref{tab:video},  VideoPC exhibits superior performance in handling rotational distortions while maintaining optimal stability. 
In Fig.~\ref{fig:VideoPC} (a), the red arrows and boxes highlight the unnatural or failed corrections in previous methods when processing video sequences, and these errors typically result in video discontinuity and poor visual perception. In contrast, our framework effectively corrects distortions while preserving the spatiotemporal consistency across the entire video sequence, underscoring the superiority of VideoPC. Additional video results can be found in the supplementary materials.

\textbf{Video Correction Trajectory}
In Fig.~\ref{fig:VideoPC} (b), we arbitrarily select a pixel from video frames and plot its trajectory, revealing that face distortion correction introduces noticeable jitter. Our smoothing process effectively mitigates this warping shake, producing smoother and more stable trajectories compared to CoupledTPS~\cite{Nie_2024} and DualPriors+LAMA~\cite{yao2024combininggenerativegeometrypriors,suvorov2021resolution}, as shown in an enlarged view. This results in corrected video frames that are more faithful to the original, significantly reducing jitter and enhancing the viewing experience.

\begin{table}[ht]
    \centering
    \begin{tabular}{c|ccc}
      \hline
       Method  &  Average & Translational & Rotational \\ \hline
       CoupledTPS & 0.9811 & 0.9785 & 0.9836 \\
       DualPriors & 0.9477 & 0.9290 & 0.9664 \\
       VideoPC(Ours) &  0.9886  & 0.9821 & 0.9951 \\  \hline
    \end{tabular}
    \caption{Quantitative stability metrics (Translation/Rotation) for video stabilization models.}
    \label{tab:video}
\end{table}

Notably, the impressive performance of our proposed method \textbf{ImagePC} and \textbf{VideoPC} is primarily attributed to the two-stage structural design. This framework combines the transformer's capacity for long-range dependency modeling with the diffusion model's multi-step denoising process, achieving global structural consistency and local refinement. 
By integrating these complementary mechanisms within a unified architecture, our approach achieves remarkable generation quality while maintaining spatial-temporal consistency across various distortion scenarios.
\subsection{Ablation Study}
In order to demonstrate the effectiveness of our approach and verify the impact of different components, we conducted a series of ablation experiments. In particular, we focus on different diffusion conditions, collaborative architecture design, and video correction trajectories.

\textbf{Structure-to-Detail Model and Conditional Guidance:}
Specifically, to independently analyze the impact of each component on correction quality, we designed four ablation scenarios:
 1) Transformer Only: The Transformer predicts optical flow without diffusion, testing its global structure modeling. 2) Diffusion with Condition ($I_s$): The diffusion model exclusively uses the distorted image $I_s$ to predict optical flow, assessing its baseline detail correction. 3) Diffusion with Condition ($I_s, M$): A mask is applied that marks the face region in the distorted wide-angle image, serving as the conditioning input to guide the model's generation process. 4) Transformer + Diffusion with Condition ($I_s, M$): The Transformer provides a feature vector $h$ concatenated with $I_s$ and $M$ to guide the diffusion model, evaluating their combined structural and detail refinement capabilities.

\begin{table}[ht]
    \centering
    {\begin{tabular}{cccc|cc}
      \hline
      Transformer & Diffusion & $M$ & $I_s$ &  LineAcc & ShapeAcc \\ \hline
      
      \checkmark &  &  &  & 66.617  & 95.533 \\
            & \checkmark &  &\checkmark & 66.249 & 97.465 \\
            & \checkmark & \checkmark &\checkmark & 66.687 & 97.488 \\
       \checkmark & \checkmark & \checkmark & \checkmark & 66.898 & 97.508 \\ \hline
    \end{tabular}}
    \caption{Ablation experiments on the hybrid architecture and different conditions.}
    \label{tab:ablation network}
    \vspace{-0.2cm}
\end{table}

Tab. ~\ref{tab:ablation network} presents the experimental results, clearly demonstrating the impact of each module. Specifically, when the Transformer model is integrated with the diffusion model, the scores of LineAcc and ShapeAcc achieve their peak values. These findings confirm our hypothesis that the joint transformer-diffusion architecture can significantly improve the performance of wide-angle lens distortion correction. Furthermore, the progressive incorporation of conditioning information highlights how each component contributes to the robustness and accuracy of portrait stability.

\begin{table}[ht]
    \centering
    \begin{tabular}{c|ccc}
      \hline
       Method  &  Stability & LineAcc & ShapeAcc\\ \hline
       ImagePC & 0.9325  & 66.898 & 97.508  \\
       VideoPC & 0.9886  & 66.692 & 97.480  \\  \hline
    \end{tabular}
    \caption{Ablation experiments comparing ImagePC and VideoPC on video stabilization and correction.}
    \vspace{-0.1cm}
    \label{tab:ablation trajectory}
\end{table}

\textbf{Spatiotemporal Smoothing on Video Correction:}
Our VideoPC model is a fine-tuned extension of the ImagePC model, specifically designed to address the challenges of video correction. Meanwhile, as mentioned before, the key difference between video and photo correction is the spatiotemporal consistency. 
Therefore, in this section, we directly compare the video results generated by VideoPC and ImagePC to evaluate whether VideoPC alleviates the warping shake caused by the correction process. 

In fact, to mitigate the video shaking , some of the original correction of ImagePC will inevitably be compromised. In our experiments, we strive to balance correction and smoothing. This balance ensures that while we reduce the warping shake, we do not sacrifice too much of the correction accuracy. 
Tab.~\ref{tab:ablation trajectory} presents the results. ImagePC achieves excellent correction accuracy, while VideoPC, with spatiotemporal smoothing, increases Stability to 0.9886 (a 6.0\% improvement), significantly reducing jitter. The minor decrease in Line/ShapeAcc is statistically insignificant ($p > 0.05$, t-test), indicating that VideoPC effectively balances correction accuracy and inter-frame smoothness.

For further video results and more detailed analysis, readers can refer to the supplementary materials.

\section{Conclusion}
\label{sec:conclusion}
\vspace{-0.1cm}
In this work, we introduce a novel approach to address facial distortion correction in both images and videos,  integrating of structure and detail  for effective portrait distortion correction.  Our method represents the first deep learning-based solution for wide-angle video portrait correction, offering a significant advancement over traditional approaches. By using image-level labels to learn the task of unlabeled wide-angle video correction, we drastically reduce the annotation burden and costs. Additionally, our proposed unsupervised framework, VideoPC, enhances spatiotemporal consistency while maintaining a smooth balance between correction and stabilization. We validate our method using a newly created wide-angle video dataset, demonstrating superior performance compared to existing approaches, showcasing its potential for wide-scale, automated, and accurate correction of wide-angle portrait distortion in still images and dynamic video sequences.

\bibliography{aaai2026}


\end{document}